\documentclass[a4paper,conference]{IEEEtran}
\IEEEoverridecommandlockouts
\usepackage{multirow}
\usepackage{acronym}
\usepackage[numbers, square, sort&compress]{natbib}
\usepackage{hyperref}
\usepackage{amsmath,amssymb,amsfonts}
\usepackage{algorithmic}
\usepackage{graphicx}
\usepackage{textcomp}
\usepackage{xcolor}
\usepackage{makecell}
\usepackage{hhline}

\def\BibTeX{{\rm B\kern-.05em{\sc i\kern-.025em b}\kern-.08em
    T\kern-.1667em\lower.7ex\hbox{E}\kern-.125emX}}
\begin{document}

\title{Multi-document Summarization: A Comparative Evaluation \\
% {\footnotesize \textsuperscript{*}Note: Sub-titles are not captured in Xplore and
% should not be used}
% \thanks{Identify applicable funding agency here. If none, delete this.}
}
%Summarization Algorithm Comparative Evaluation for Multi-document Use Case
\author{\IEEEauthorblockN{Kushan Hewapathirana, Nisansa de Silva}
\IEEEauthorblockA{\textit{Department of Computer Science \& Engineering} \\
\textit{University of Moratuwa, Sri Lanka}\\
\texttt{\{kushan.22,nisansa\}@cse.mrt.ac.lk}}
%\and
%\IEEEauthorblockN{Nisansa de Silva}
%\IEEEauthorblockA{\textit{Dept. of Computer Science \& Engineering} \\
%\textit{University of Moratuwa, Sri Lanka}\\
%\texttt{NisansaDdS@cse.mrt.ac.lk}}
\and
\IEEEauthorblockN{C.D. Athuraliya}
\IEEEauthorblockA{\textit{ConscientAI, Sri Lanka} \\
\texttt{cd@conscient.ai}}
% \and
% \IEEEauthorblockN{4\textsuperscript{th} Given Name Surname}
% \IEEEauthorblockA{\textit{dept. name of organization (of Aff.)} \\
% \textit{name of organization (of Aff.)}\\
% City, Country \\
% email address or ORCID}
% \and
% \IEEEauthorblockN{5\textsuperscript{th} Given Name Surname}
% \IEEEauthorblockA{\textit{dept. name of organization (of Aff.)} \\
% \textit{name of organization (of Aff.)}\\
% City, Country \\
% email address or ORCID}
% \and
% \IEEEauthorblockN{6\textsuperscript{th} Given Name Surname}
% \IEEEauthorblockA{\textit{dept. name of organization (of Aff.)} \\
% \textit{name of organization (of Aff.)}\\
% City, Country \\
% email address or ORCID}
}

%\author{\IEEEauthorblockN{Kushan Hewapathirana}
%\IEEEauthorblockA{\textit{Department of Computer Science \& Engineering} \\
%\textit{University of Moratuwa, Sri Lanka}\\
%\texttt{kushan.22@cse.mrt.ac.lk}}\\
%\IEEEauthorblockA{\textit{ConscientAI, Sri Lanka} \\
%\texttt{cd@conscient.ai}}
%\and
%\IEEEauthorblockN{Nisansa de Silva}
%\IEEEauthorblockA{\textit{Department of Computer Science \& Engineering} \\
%\textit{University of Moratuwa, Sri Lanka}\\
%\texttt{nisansa@cse.mrt.ac.lk}}
%\and
%\IEEEauthorblockN{C.D. Athuraliya}
%\IEEEauthorblockA{\textit{ConscientAI, Sri Lanka} %\\
%\texttt{cd@conscient.ai}}
%}

\DeclareRobustCommand{\IEEEauthorrefNumMark}[1]{\smash{\textsuperscript{\footnotesize #1}}}

\author{\IEEEauthorblockN{Kushan Hewapathirana\IEEEauthorrefmark{1}\IEEEauthorrefNumMark{1}\textsuperscript{,}\IEEEauthorrefNumMark{2}, Nisansa de Silva\IEEEauthorrefmark{2}\IEEEauthorrefNumMark{1}, C.D. Athuraliya\IEEEauthorrefmark{3}\IEEEauthorrefNumMark{2}}
\IEEEauthorblockA{\IEEEauthorrefNumMark{1}\textit{Department of Computer Science \& Engineering, University of Moratuwa, Sri Lanka}\\
\texttt{\{\IEEEauthorrefmark{1}kushan.22,\IEEEauthorrefmark{2}nisansa\}@cse.mrt.ac.lk}}
\IEEEauthorblockA{\IEEEauthorrefNumMark{2}\textit{ConscientAI, Sri Lanka} \\
\texttt{\IEEEauthorrefmark{3}cd@conscient.ai}}
}

% \IEEEoverridecommandlockouts
% \IEEEpubid{\makebox[\columnwidth]{ 979-8-3503-2363-4/23/\$31.00~\copyright~2023~IEEE \hfill} \hspace{\columnsep}\makebox[\columnwidth]{ }}

\IEEEoverridecommandlockouts
\IEEEpubid{\begin{minipage}[t]{\textwidth}\ \\[10pt]
        \normalsize{979-8-3503-2363-4/23/\$31.00 \copyright 2023 IEEE}
\end{minipage}} 

% \IEEEoverridecommandlockouts
% \IEEEpubid{\makebox[\columnwidth]{979-8-3503-2363-4/23/\$31.00~\copyright2023 IEEE \hfill}
% \hspace{\columnsep}\makebox[\columnwidth]{ }}

\maketitle

\begin{abstract}
% This document is a model and instructions for \LaTeX.
% This and the IEEEtran.cls file define the components of your paper [title, text, heads, etc.]. *CRITICAL: Do Not Use Symbols, Special Characters, Footnotes, 
% or Math in Paper Title or Abstract.
This paper is aimed at evaluating state-of-the-art models for Multi-document Summarization (MDS) on different types of datasets in various domains and investigating the limitations of existing models to determine future research directions. To address this gap, we conducted an extensive literature review to identify state-of-the-art models and datasets. We analyzed the performance of PRIMERA and PEGASUS models on BigSurvey-MDS and MSˆ2 datasets, which posed unique challenges due to their varied domains. Our findings show that the General-Purpose Pre-trained Model LED outperforms PRIMERA and PEGASUS on the MSˆ2 dataset. We used the ROUGE score as a performance metric to evaluate the identified models on different datasets. Our study provides valuable insights into the models' strengths and weaknesses, as well as their applicability in different domains. This work serves as a reference for future MDS research and contributes to the development of accurate and robust models which can be utilized on demanding datasets with academically and/or scientifically complex data as well as generalized, relatively simple datasets.

\end{abstract}

\renewcommand\IEEEkeywordsname{Keywords}

\section{Introduction}
Despite the maturity of single-document summarization, MDS remains to be a challenging Natural Language Processing (NLP) task because it involves combining information from multiple sources, often with conflicting, duplicate or complementary information, in order to generate a summary that is representative of the overall content ~\cite{ma2020multi,afsharizadeh2022survey,abid2022multi}.
% MDS is a NLP task that generates a summary of multiple often related documents~\cite{ma2020multi,afsharizadeh2022survey,abid2022multi}. 
The goal of MDS is to condense a collection of documents into a single, cohesive summary that captures the main points and ideas of the original documents.
Automatic summarization, be it single-document or multi-document, can be divided into two primary categories: extractive and abstractive~\cite{pasunuru2021efficiently,abid2022multi,ma2020multi,afsharizadeh2022survey,ma2021dependency,wolhandler2022multi, kumarasinghe2022automatic}.

\begin{itemize}
\item \textit{Extractive text summaries} contain keywords, phrases, and sentences that are extracted \textit{verbatim} from the source documents~\cite{ma2020multi,afsharizadeh2022survey,pasunuru2021efficiently}. 
% Fix summaries to summary
\item \textit{Abstractive text summaries} generate informative summaries, including paraphrased sentences and new terms that might not be found in the original documents~\cite{ma2020multi,afsharizadeh2022survey,pasunuru2021efficiently,abid2022multi, kumarasinghe2022automatic}.
\end{itemize}

In MDS, the documents that are being summarized can be of different types. Some common types of documents used in MDS include~\cite{ma2020multi,afsharizadeh2022survey}:
\begin{itemize}
\item \textit{Short sources}: These are short documents, such as tweets, product reviews or headlines, that convey a smaller amount of information. The overall quantity of the input data, on the other hand, is large~\cite{ma2020multi,afsharizadeh2022survey, abid2022multi}. 
\item \textit{Long sources}: These are lengthy documents, such as news articles or research papers, which contain a large amount of information and detail. The quantity of the input data is generally small~\cite{ma2020multi,afsharizadeh2022survey,pasunuru2021efficiently, yu2022evaluating,kumarasinghe2022automatic}.
\item \textit{Hybrid sources}: Contains one or few long documents with several to many short documents (e.g., A scientific summary from a long paper with several short corresponding citations)~\cite{ma2020multi,afsharizadeh2022survey,yu2022evaluating}. 
\end{itemize}

It is noteworthy that each type of source has its own specific characteristics and challenges; thus the summarization techniques used on them differ accordingly~\cite{ma2021dependency,lakshmihybrid,ma2020multi,wolhandler2022multi}. Therefore, it is important to select the appropriate summarization technique for each type of source to achieve desired results.
%
% Moreover, the type of documents used in multi-document summarization can have an impact on the quality of the summary, as different types of documents may require different summarization techniques to extract the most relevant information~\cite{lakshmihybrid,ma2020multi,wolhandler2022multi}.
For example, long sources may require more sophisticated techniques to identify the most important sentences, while short sources may be more easily summarized by simply extracting the keywords or phrases~\cite{wolhandler2022multi,abid2022multi}.

There is a significant research gap in evaluating state-of-the-art models on recently released datasets, particularly their performance when exposed to different domains, including long, short, and hybrid data~\cite{afsharizadeh2022survey,ma2020multi, wolhandler2022multi,deyoung2023multi}. This suggests the importance of investigating the limitations of state-of-the-art techniques and identifying future directions.
% There is a significant gap in research regarding the evaluation of state-of-the-art models on recently released datasets, particularly in relation to their performance when exposed to different domains, including long, short, and hybrid data~\cite{afsharizadeh2022survey,ma2020multi, wolhandler2022multi,deyoung2023multi}. This suggests the importance of investigating the limitations of state-of-the-art techniques and identifying future directions.

To address this gap, an extensive literature review was conducted to identify state-of-the-art MDS techniques and applicable datasets. However, some of these baseline models have not yet been evaluated against each other with academically and/or scientifically complex datasets with domain-specific content, which limits our understanding of their performance on such recently-released, complex, and diverse datasets.

This research aims to evaluate the performance of state-of-the-art models in the context of commonly-used datasets. Our objective is to identify the capabilities and limitations of these models and their effectiveness in various applications across domains. This work can serve as a reference for future research in this area, thus leading to more accurate and robust models that can handle a variety of datasets as well as useful downstream applications.

\section{Background and related work}
This section presents an overview of the models, datasets, and techniques employed in this study. It highlights the work carried out by others in various areas relevant to this study.

\subsection{Pre-trained Models in Summarization Tasks}
% Recently, pre-trained models have shown great potential in various NLP tasks, including MDS. 
% In this section, we discuss pre-trained models used in MDS tasks.
%
BERTSUM~\cite{liu2019text}, BART~\cite{lewis2019bart}, PEGASUS~\cite{zhang2020pegasus}, and T5~\cite{raffel2020exploring} are some pre-trained models that have been widely used in MDS tasks. 
%
% Although PEGASUS~\cite{zhang2020pegasus} and T5~\cite{raffel2020exploring} were not initially designed for MDS, they have still been utilized for this task.  
%
Additionally, transformer-based architectures such as Longformer~\cite{beltagy2020longformer} and BigBird~\cite{zaheer2020big} have been successful in MDS due to their ability to process long sequences. CDLM~\cite{caciularu2021cdlm} is a recent follow-up work that pre-trains Longformer~\cite{beltagy2020longformer} for cross-document tasks but only addresses encoder-specific tasks and is not suitable for text generation.

PRIMERA~\cite{xiao2021primer} is a pre-training method proposed for neural MDS based on the pre-training of the LED architecture~\cite{beltagy2020longformer}. The PRIMERA method is trained on NewSHead, a corpus of $369,940$ news clusters with similar topics~\cite{xiao2021primer}. To create synthetic summaries for pre-training, the authors adopt the \textit{Entity Pyramid} strategy which is used to select the most important sentences in a document. The Entity Pyramid strategy is based on the frequency of entities that appear in the document. \citet{xiao2021primer} demonstrate the effectiveness of the PRIMERA method for MDS compared to previous methods.

DAMEN~\cite{moro2022discriminative} is a novel MDS method proposed for the medical domain. The authors argue that current state-of-the-art Transformer-based solutions for MDS are not suitable for the medical domain because they either truncate inputs or fail to distinguish between relevant and irrelevant information. DAMEN involves using two BERT models, known as the \textit{Indexer} and the \textit{Discriminator}. The \textit{Indexer} encodes the background information and documents in a cluster, resulting in dense embedding representations. The \textit{Discriminator} then selects the top $k$ documents through a comparison with the background. The background is then combined with the retrieved documents and passed to BART, a probabilistic neural method, to generate the multi-document summary.

CGSUM~\cite{chen2022comparative} is a recent approach for scientific paper summarization. The authors constructed a comparative scientific summarization corpus (CSSC) and presented a comparative citation-guided summarization (CGSUM) method. The method involves extracting citation sentences and using them to compare and rank different scientific papers~\cite{chen2022comparative}. The top-ranked papers are then used to generate a summary of the target paper. The authors demonstrate the effectiveness of CGSUM on several scientific papers compared to previous methods~\cite{chen2022comparative}.

\subsection{Commonly-used datasets in MDS}

 % In recent years, various datasets have been published to evaluate and train MDS models. 
 This section provides an overview of commonly-used datasets in text summarization literature.
 
\subsubsection{DUC and TAC Datasets}
% \textbf{\textit{DUC and TAC Datasets}}
 
 The Document Understanding Conference (DUC)\footnote{\url{https://duc.nist.gov/}} and the Text Analysis Conference (TAC)\footnote{\url{https://tac.nist.gov/}} are two events that have contributed significantly to the development of text summarization research. From 2001 to 2007, DUC held annual text summarization competitions, which published datasets for researchers to evaluate models. In 2008, DUC changed its name to TAC but continued to publish datasets. The DUC and TAC datasets primarily consist of news articles from various domains, including politics, natural disasters, and biographies. Although useful for model evaluation, these datasets have certain limitations, such as being relatively small and biased towards the first sentence of news articles~\cite{afsharizadeh2022survey,ma2020multi}.
 
 % \subsubsection{WikiSum Dataset}
 % The WikiSum dataset is a widely used dataset in abstractive multi-document summarization research~\cite{afsharizadeh2022survey, ma2020multi, xiao2021primer}. The dataset is created using Wikipedia articles and their cited sources or the top-10 results from a Google search using the Wikipedia article's theme as the query~\cite{liu2018generating}. The golden summaries are the actual Wikipedia articles, but some of the URLs were not available, and some source documents contained duplicate information~\cite{ma2020multi}. To address these issues, the dataset has been cleaned by \citet{liu2019hierarchical}.
 
 % \subsubsection{Multi-News Dataset}
 % The Multi-News dataset is a large-scale dataset in the news domain, consisting of over 56,000 article-summary pairs~\cite{fabbri2019multi}. The articles and human-written summaries are sourced from over 1,500 websites\footnote{\url{http://newser.com/}} and come with trace-back links to the original documents. This dataset is unique as it is the first large-scale dataset for multi-document summarization in the news domain, with contributions from 20 editors~\cite{fabbri2019multi}. The dataset presents an opportunity for models to learn from a variety of source documents and summaries, making it a valuable resource for multi-document summarization research~\cite{ma2020multi}.

% Nisansa updated up to here
\begin{table*}[htbp]
\centering
\caption{Summary of datasets used for evaluation}
\label{tab:dataset}
\begin{tabular}{|l|c|c|l|}
\hline
\textbf{Dataset} &
  \makecell{\textbf{Total number}\\ \textbf{of documents}} &
  \makecell{\textbf{Average number of} \\ \textbf{documents per cluster}} &
  \multicolumn{1}{c|}{\textbf{Domain}} \\ \hline
\textbf{Multi-News~\cite{fabbri2019multi}}              & 56K~{\footnotesize\cite{xiao2021primer}}   & 3.5~{\footnotesize\cite{xiao2021primer}}  & News articles~\cite{fabbri2019multi}                                    \\ \hline
\textbf{Multi-Xscience~\cite{lu2020multi}}          & 40K~{\footnotesize\cite{xiao2021primer}}   & 2.8~{\footnotesize\cite{xiao2021primer}}  & Related-work section   in scientific articles~\cite{lu2020multi}    \\ \hline
\textbf{Wikisum~\cite{liu2018generating}}                 & 1.5M~{\footnotesize\cite{xiao2021primer}}  & 40~{\footnotesize\cite{xiao2021primer}}   & Wikipedia articles~\cite{liu2018generating}                               \\ \hline
\textbf{BigSurvey-MDS~\cite{liu2023generating}}           & 430K~{\footnotesize\cite{xiao2021primer}}  & 61.4~{\footnotesize\cite{xiao2021primer}} & Human-written survey   papers on various domains~\cite{liu2023generating} \\ \hline
% \textbf{PEERSUM~\cite{li2022peersum}}                 & 11.9K~\cite{li2022peersum} & 7.75~\cite{li2022peersum} & Peer reviews of   scientific publications~\cite{li2022peersum}        \\ \hline
\textbf{MS\textasciicircum{}2~{\footnotesize\cite{deyoung2021ms2}}} &
  470K~{\footnotesize\cite{deyoung2021ms2}} &
  23.5~{\footnotesize\cite{deyoung2021ms2}} &
  Reviews of scientific   publications in medical domain~{\footnotesize\cite{deyoung2021ms2}} \\ \hline
\textbf{Rotten Tomato   Dataset~\cite{deyoung2023multi}} & 244K~\cite{deyoung2023multi}  & 26.8~\cite{deyoung2023multi} & Movie reviews~\cite{deyoung2023multi}                                    \\ \hline
\end{tabular}
\end{table*}

 %\textbf{\textit{WikiHow Dataset}}
\subsubsection{WikiHow Dataset}

 The WikiHow dataset is a large-scale dataset consisting of over $230,000$ article-summary pairs extracted from an online knowledge base~\cite{koupaee2018wikihow}. The articles in this dataset cover a wide range of topics and are written by different authors, resulting a diverse representation of writing styles. This sets WikiHow apart from other summarization datasets, which are limited to news articles and the journalistic writing style~\cite{koupaee2018wikihow}. The evaluation of existing summarization methods on the WikiHow dataset provides insights into the challenges and limitations of existing datasets, and sets a baseline for further improvement~\cite{ma2020multi,koupaee2018wikihow,grusky2018newsroom,fabbri2019multi}.
 
%  \subsubsection{Rotten Tomatoes Dataset}
 
% The Rotten Tomatoes dataset comprises of movie reviews and meta-reviews constructed from professional critics and user comments~\cite{deyoung2023multi}. The meta-reviews, which summarize the constituent input reviews, are synthesized by professional editors and are associated with a numerical "Tomatometer\footnote{\url{http://rottentomatoes.com/}}" score that reflects the aggregate critic reception of a film. The dataset provides a valuable resource for sentiment analysis and summarization research in the movie review domain. A BERT model has been trained to measure sentiment in the movie reviews, showing strong correlation with the "Tomatometer" score~\cite{deyoung2023multi}.

\section{Methodology}
This section presents the methodology employed in the research conducted. Each component of the methodology is detailed under each subsections below.
\subsection{Evaluated models}
We carried out an extensive literature review to identify state-of-the-art models for MDS. The model selection process involved considering the overall performance of the models and their ROUGE (Recall-Oriented Understudy for Gisting Evaluation)~\cite{lin2004rouge} scores. The publication year and venue were also taken into account during model selection.

In this study, we evaluated the performance of three summarization models: PRIMERA~\cite{xiao2021primer}, PEGASUS~\cite{zhang2020pegasus}, and LED~\cite{beltagy2020longformer}. PRIMERA has outperformed many other models and baselines in previous studies~\cite{xiao2021primer, afsharizadeh2022survey, ma2020multi, deyoung2023multi, wolhandler2022multi}. However, when considering the sentiment focus of the models on the Rotten Tomatoes dataset~\cite{Leon2020Rotten}, PEGASUS outperforms PRIMERA~\cite{deyoung2023multi}. On the other hand, LED is a pre-trained model that is commonly used as a baseline in related literature~\cite{afsharizadeh2022survey, ma2020multi, deyoung2023multi, wolhandler2022multi}. Moreover, PRIMERA is a modified version of the LED architecture, which is why we selected these three models for our analysis~\cite{xiao2021primer}.

\citet{beltagy2020longformer} proposes a Longformer-based model for seq2seq learning that has both the encoder and decoder Transformer stacks. Unlike the full self-attention pattern used in the Longformer encoder, \citet{beltagy2020longformer} uses an efficient local+global attention pattern. Scalability of the LED is linear with respect to the input size~\cite{beltagy2020longformer}. However, pre-training LED is expensive, so the authors initialize LED parameters from BART and follow BART's architecture~\cite{liu2019text} in terms of number of hidden layers~\cite{beltagy2020longformer}. To process longer inputs, LED extends the positional embeddings to $16$K tokens and initializes the new position embedding matrix by repeatedly copying BART's $1$K position embeddings $16$ times~\cite{beltagy2020longformer, liu2019text}. LED comes in two sizes, LED-base and LED-large, with $6$ and $12$ layers in both encoder and decoder stacks, respectively~\cite{beltagy2020longformer}.
PEGASUS and PRIMERA use Gap Sentence Generation (GSG) as a pretraining objective~\cite{xiao2021primer}. GSG involves masking certain sentences in the input and training the model to generate them. PEGASUS uses a standard approach to GSG, whereas PRIMERA introduces a new masking strategy specifically designed for MDS~\cite{xiao2021primer, zhang2020pegasus}.

In PRIMERA's approach to GSG, authors select and mask \textit{m} summary-like sentences from the input documents to be summarized. Each selected sentence is replaced by a single token \textit{[sent-mask]} in the input, and the model is trained to generate the concatenation of these sentences as a \textit{pseudo-summary}~\cite{xiao2021primer}. This approach closely resembles abstractive summarization, as the model must use the information from the rest of the documents to reconstruct masked sentences. The main difference from standard GSG is PRIMERA's focus on selecting the sentences that best summarize or represent a set of related input documents, which can be considered as a \textit{cluster}, instead of just a single document~\cite{xiao2021primer, zhang2020pegasus}. This novel approach for GSG sentence masking is called Entity Pyramid. This method leverages the frequency of entities appearing in the input documents to select representative sentences for masking~\cite{xiao2021primer}. Specifically, sentences that contain entities with a frequency greater than one are chosen and evaluated using Cluster ROUGE as the criterion. For instance, if sentence ten in document two is the most representative sentence for entity one, it is masked by replacing it with a mask token~\cite{xiao2021primer}.

\subsection{Datasets used for evaluation}
The selection of datasets for this study involved considering the popularity and novelty of the dataset. Datasets from different domains were selected, and the data type (i.e. short, long, or hybrid documents) was also taken into consideration. A variety of datasets were selected based on the aforementioned criteria (Table~\ref{tab:dataset}).

The Multi-News dataset~\cite{fabbri2019multi} is a collection of news articles and their corresponding human-written summaries from the website newser.com\footnote{\url{https://www.newser.com/}}. The summaries are created by professional editors and include links to the original articles~\cite{fabbri2019multi}. The dataset is available through Wayback-archived links and scripts that allow for easy reproduction of the data. It is the first large-scale dataset for MDS of news articles~\cite{fabbri2019multi}. The dataset is sourced from over $1,500$ news sites, making it more diverse than previous datasets such as DUC, which come from only two sources. On the other hand, the Newsroom dataset~\cite{grusky2018newsroom} covers only $38$ news sources~\cite{fabbri2019multi}. A small group of $20$ editors contributed to $85$\% of the summaries in the Multi-News dataset, providing a diverse range of perspectives for summarizing news articles~\cite{fabbri2019multi}.

The Multi-XScience dataset~\cite{lu2020multi} is created by combining arXiv.org\footnote{\url{https://arxiv.org/}} papers and Microsoft Academic Graph (MAG)~\cite{sinha2015overview} to form pairs of target summaries and multi-reference documents. The authors took care to maximize its usefulness by cleaning the \LaTeX ~source of $1.3$ million arXiv papers, aligning them with their references in MAG, and five cleaning iterations followed by human verification~\cite{lu2020multi}. Multi-XScience has $60$\% more references than Multi-News, making it suitable for the MDS setting. Despite being smaller than WikiSum, Multi-XScience is better suited for abstractive summarization because its reference summaries contain more novel n-grams than the source~\cite{lu2020multi,liu2018generating}. The dataset has a high novel n-grams score, which means it has less extractive bias, resulting in better abstraction for summarization models~\cite{lu2020multi}. However, the high level of abstractiveness makes the dataset challenging because models cannot extract sentences from the reference articles~\cite{lu2020multi}.

The authors of the WikiSum dataset~\cite{liu2018generating} proposed using Wikipedia as a collection of summaries on various topics. They extract source material to be summarized from two subsets of all documents: 1) Cited sources: they extract text without markup from the crawlable citation documents for each article. 2) Web Search results: they crawl the search results from the Google search engine using the article section titles as queries, remove the Wikipedia article itself, and extract only the text. They supplement the source documents with web search results since many articles have few citations. The dataset is two orders-of-magnitude larger than previous summarization datasets in terms of the total words. 
% To ensure consistent train/development/test data across corpus-comparison experiments, the authors restrict the articles to those with at least one crawlable citation and divide them roughly into $80/10/10$ for train/development/test subsets, resulting in $1865750$, $233252$, and $232998$ examples, respectively. 
The WikiSum dataset is suitable for summarization tasks and provides a rich source of information~\cite{liu2018generating}.

Current datasets for summarization typically focus on either producing structureless summaries covering a few input documents or on summarizing a single document into a multi-section summary~\cite{ma2020multi,liu2023generating}. However, these datasets and methods do not meet the requirements of summarizing multiple academic papers into a structured summary. To address this issue, BigSurvey was proposed, which is a large-scale dataset for generating comprehensive summaries of academic papers on each topic~\cite{liu2023generating}.

The authors collected target summaries from more than $7,000$ survey papers and used the abstracts of their $430,000$ citations as input documents. BigSurvey contains two-level target summaries for multiple academic papers on the same topic. The long summary aims to comprehensively cover the salient content of the reference papers in different aspects, while the shorter summary is more concise and serves as a summary of the long summary. The authors built two subsets of the dataset: BigSurvey-MDS and BigSurveyAbs~\cite{liu2023generating}.
% The authors collected target summaries from more than 7,000 survey papers and used the abstracts of their 430,000 citations as input documents. BigSurvey contains two-level target summaries for dozens of academic papers on the same topic. The long summary aims to comprehensively cover the salient content of the reference papers in different aspects, while the shorter summary is more concise and serves as a summary of the long summary. The authors build two subsets of the dataset: BigSurvey-MDS and BigSurveyAbs~\cite{liu2023generating}. 

In this research, we use the BigSurvey-MDS dataset, as it covers the salient content of the reference papers in different aspects.
Each instance in the BigSurvey-MDS dataset corresponds to one survey paper from \hyperlink{https://arxiv.org/}{arXiv.org}, which typically has tens or hundreds of reference papers. Due to copyright issues, BigSurvey-MDS does not include the body sections of these reference papers, but uses their abstracts as input documents. The abstracts can be regarded as summaries written by the authors of the reference papers, which include the papers' salient information. For each survey paper, the authors collect up to $200$ reference paper abstracts and truncate them to no more than $200$ words, which are then used as input documents of the BigSurvey-MDS~\cite{liu2023generating}. 

The authors classified sentences in the introduction section of the survey paper into three sections and used them to compose the structured summary as the target in each example of the BigSurvey-MDS. The objective, result, and other types of content were merged into the section named as "other", since they appear less frequently than the background and method in the survey papers' introduction section. To prepare these three sections in the target summary, the authors first collected the introduction section from a survey paper. If there is no introduction section, they extracted the first 1,024 words after the abstract part. Then, they classified the sentences and concatenated those classified as the same type to form the three sections in the target summary. The authors filtered out the examples with too short input sequences or target summaries~\cite{liu2023generating}.

In addition to the BigSurvey-MDS dataset, we employed the MSˆ2 dataset to evaluate the performance of state-of-the-art models in the biomedical domain. This dataset comprises more than $470,000$ documents and $20,000$ summaries drawn from scientific literature, and was specifically designed to support the development of systems that can assess and consolidate conflicting evidence from multiple studies. It is the first publicly available, large-scale, MDS dataset in the biomedical domain. Notably, each review in the MSˆ2 dataset summarizes an average of $23$ studies, and the input documents contain contradictory evidence, setting it apart from other datasets~\cite{deyoung2021ms2}.

The Rotten Tomatoes (RT) dataset~\cite{Leon2020Rotten} comprises of movie reviews and meta-reviews constructed from professional critics and user comments. The meta-reviews, which summarize the constituent input reviews, are synthesized by professional editors and are associated with a numerical \textit{Tomatometer}\footnote{\url{http://rottentomatoes.com/}} score that reflects the aggregate critic reception of a movie. The dataset provides a valuable resource for sentiment analysis and summarization research in the movie review domain. A BERT model has been trained to measure sentiment in movie reviews, showing a strong correlation with the "Tomatometer" score~\cite{deyoung2023multi}. This stands in contrast with the earlier single docu- ment sentiment analysis datasets for RT~\cite{pang2005seeing} or otherwise~\cite{gunathilaka2022aspect}.

\subsection{Collection of results and experiments}
We collected reported results from other studies that used the same models and datasets. The ROUGE score~\cite{lin2004rouge} was used as the metric of performance, and in cases where conflicting results were found, the latest version of the study and the conference where it was published were considered. The original parameters used in the studies introducing the models were adhered to.

% If any of the selected datasets were not previously tested using the selected models, the research team ran the models on the dataset and reported the results. The results are summarized in a table for easy comparison and analysis.
In this study, we evaluated the performance of the PRIMERA\footnote{PRIMERA code - \url{https://github.com/allenai/PRIMER}}~\cite{xiao2021primer} model on the recently introduced BigSurvey-MDS\footnote{BigSurvey-MDS dataset - \url{https://github.com/StevenLau6/BigSurvey}} ~\cite{liu2023generating}. For this, we used the same parameters as the original PRIMERA setup and ran the model on the dataset's test set. The results obtained were then reported and summarized in Table~\ref{tab:ROUGE-scores} to facilitate comparison and analysis. As per our understanding, this is the first evaluation of the PRIMERA model on the BigSurvey MDS dataset.

\subsection{Evaluation metrics}
Evaluation of MDS models is a critical aspect in the development and testing of them~\cite{ma2020multi}.
% In this work, several evaluation metrics were used to assess the performance of MDS models. Among these, ROUGE~\cite{lin2004rouge} is one of the most commonly used evaluation metrics for multi-document summarization. 
ROUGE~\cite{lin2004rouge}, 
%one of the most commonly used evaluation metrics, 
is based on measuring the overlap between the generated summary and the reference summary, and it calculates a score based on the recall and precision of the generated summary~\cite{lin2004rouge}. The use of ROUGE has been reported in several studies for evaluating MDS models~\cite{afsharizadeh2022survey,ma2020multi}.
ROUGE has different variants that include ROUGE-N and ROUGE-L~\cite{lin2004rouge}. ROUGE-N measures the overlap of \textit{n}-grams between the reference and candidate summaries. On the other hand, ROUGE-L calculates the similarity at the sentence level using the longest common subsequence~\cite{lin2004rouge}. % It calculates the length of the longest sequence of words that appear in both the generated summary and the reference summary, divided by the total number of words in the reference summary~\cite{lin2004rouge}.

\begin{table*}[!htbp]
\centering
\caption{ROUGE scores of different models on different datasets}
\label{tab:ROUGE-scores}
\begin{center}
\begin{tabular}{|c|c|ccc|}
\hline
 \multirow{2}{*}{\textbf{Datasets}} & \multirow{2}{*}{\textbf{Metric}} & \multicolumn{3}{c|}{\textbf{Models}} \\ 
 \hhline{~~---}
                            &     & \multicolumn{1}{c|}{PRIMERA~\cite{xiao2021primer}} & \multicolumn{1}{c|}{PEGASUS~\cite{zhang2020pegasus}} & \multicolumn{1}{c|}{LED~\cite{beltagy2020longformer}}  \\ \hline
\multirow{3}{*}{\textbf{Multi-News~\cite{fabbri2019multi}}}                & ROUGE-1 & \multicolumn{1}{c|}{\textbf{42.0}~{\footnotesize\cite{xiao2021primer}}}    & \multicolumn{1}{c|}{32.0~{\footnotesize\cite{xiao2021primer}}}    & \multicolumn{1}{c|}{17.3~{\footnotesize\cite{xiao2021primer}}} \\ \cline{2-5} 
                                                    & ROUGE-2 & \multicolumn{1}{c|}{\textbf{13.6}~{\footnotesize\cite{xiao2021primer}}}    & \multicolumn{1}{c|}{10.1~{\footnotesize\cite{xiao2021primer}}}    & \multicolumn{1}{c|}{3.7~{\footnotesize\cite{xiao2021primer}}}  \\ \cline{2-5} 
                                                    & ROUGE-L & \multicolumn{1}{c|}{\textbf{20.8}~{\footnotesize\cite{xiao2021primer}}}    & \multicolumn{1}{c|}{16.7~{\footnotesize\cite{xiao2021primer}}}    & \multicolumn{1}{c|}{10.4~{\footnotesize\cite{xiao2021primer}}} \\ \hline
\multirow{3}{*}{\textbf{Multi-XScience~\cite{lu2020multi}}}                & ROUGE-1 & \multicolumn{1}{c|}{\textbf{29.1}~{\footnotesize\cite{xiao2021primer}}}    & \multicolumn{1}{c|}{27.6~{\footnotesize\cite{xiao2021primer}}}    & \multicolumn{1}{c|}{14.6~{\footnotesize\cite{xiao2021primer}}} \\ \cline{2-5} 
                                                    & ROUGE-2 & \multicolumn{1}{c|}{\textbf{4.6}~{\footnotesize\cite{xiao2021primer}}}     & \multicolumn{1}{c|}{\textbf{4.6}~{\footnotesize\cite{xiao2021primer}}}     & \multicolumn{1}{c|}{1.9~{\footnotesize\cite{xiao2021primer}}}  \\ \cline{2-5} 
                                                    & ROUGE-L & \multicolumn{1}{c|}{\textbf{15.7}~{\footnotesize\cite{xiao2021primer}}}    & \multicolumn{1}{c|}{15.3~{\footnotesize\cite{xiao2021primer}}}    & \multicolumn{1}{c|}{9.9~{\footnotesize\cite{xiao2021primer}}}  \\ \hline
\multirow{3}{*}{\textbf{WikiSum~\cite{liu2018generating}}}                   & ROUGE-1 & \multicolumn{1}{c|}{\textbf{28.0}~{\footnotesize\cite{xiao2021primer}}}    & \multicolumn{1}{c|}{24.6~{\footnotesize\cite{xiao2021primer}}}    & \multicolumn{1}{c|}{10.5~{\footnotesize\cite{xiao2021primer}}} \\ \cline{2-5} 
                                                    & ROUGE-2 & \multicolumn{1}{c|}{\textbf{8.0}~{\footnotesize\cite{xiao2021primer}}}     & \multicolumn{1}{c|}{5.5~{\footnotesize\cite{xiao2021primer}}}     & \multicolumn{1}{c|}{2.4~{\footnotesize\cite{xiao2021primer}}}  \\ \cline{2-5} 
                                                    & ROUGE-L & \multicolumn{1}{c|}{\textbf{18.0}~{\footnotesize\cite{xiao2021primer}}}    & \multicolumn{1}{c|}{15.0~{\footnotesize\cite{xiao2021primer}}}    & \multicolumn{1}{c|}{8.6~{\footnotesize\cite{xiao2021primer}}}  \\ \hline
\multirow{3}{*}{\textbf{BigSurvey-MDS~\cite{liu2023generating}}}             & ROUGE-1 & \multicolumn{1}{c|}{23.9}       & \multicolumn{1}{c|}{\textbf{38.9}~{\footnotesize\cite{liu2023generating}}}    & \multicolumn{1}{c|}{39.8~{\footnotesize\cite{liu2023generating}}} \\ \cline{2-5} 
                                                    & ROUGE-2 & \multicolumn{1}{c|}{4.1}       & \multicolumn{1}{c|}{9.0~{\footnotesize\cite{liu2023generating}}}     & \multicolumn{1}{c|}{\textbf{9.4}~{\footnotesize\cite{liu2023generating}}}  \\ \cline{2-5} 
                                                    & ROUGE-L & \multicolumn{1}{c|}{11.7}       & \multicolumn{1}{c|}{\textbf{16.2}~{\footnotesize\cite{liu2023generating}}}    & \multicolumn{1}{c|}{16.1~{\footnotesize\cite{liu2023generating}}} \\ \hline
% \multirow{3}{*}{\textbf{PEERSUM~\cite{li2022peersum}}}                   & ROUGE-1 & \multicolumn{1}{c|}{-}       & \multicolumn{1}{c|}{43.7~{\footnotesize\cite{li2022peersum}}}    & \multicolumn{1}{c|}{-}    \\ \cline{2-5} 
%                                                     & ROUGE-2 & \multicolumn{1}{c|}{-}       & \multicolumn{1}{c|}{18.8~{\footnotesize\cite{li2022peersum}}}    & \multicolumn{1}{c|}{-}    \\ \cline{2-5} 
%                                                     & ROUGE-L & \multicolumn{1}{c|}{-}       & \multicolumn{1}{c|}{31.4~{\footnotesize\cite{li2022peersum}}}    & \multicolumn{1}{c|}{-}    \\ \hline
\multirow{3}{*}{\textbf{MSˆ2~{\footnotesize\cite{deyoung2021ms2}}}}                      & ROUGE-1 & \multicolumn{1}{c|}{12.8}       & \multicolumn{1}{c|}{12.7}       & \textbf{25.8}~{\footnotesize\cite{wang2022overview}} \\ \cline{2-5} 
                                                    & ROUGE-2 & \multicolumn{1}{c|}{2.0}       & \multicolumn{1}{c|}{1.5}       & \textbf{8.4}~{\footnotesize\cite{wang2022overview}}  \\ \cline{2-5} 
                                                    & ROUGE-L & \multicolumn{1}{c|}{8.1}       & \multicolumn{1}{c|}{8.3}       & \textbf{19.3}~{\footnotesize\cite{wang2022overview}} \\ \hline
\multirow{3}{*}{\textbf{Rotten Tomatoes   Dataset~\cite{Leon2020Rotten}}} & ROUGE-1 & \multicolumn{1}{c|}{25.4~{\footnotesize\cite{deyoung2023multi}}}    & \multicolumn{1}{c|}{\textbf{27.4}~{\footnotesize\cite{deyoung2023multi}}}    & \multicolumn{1}{c|}{25.6~{\footnotesize\cite{deyoung2023multi}}} \\ \cline{2-5} 
                                                    & ROUGE-2 & \multicolumn{1}{c|}{8.4~{\footnotesize\cite{deyoung2023multi}}}     & \multicolumn{1}{c|}{\textbf{9.5}~{\footnotesize\cite{deyoung2023multi}}}     & 8.0~{\footnotesize\cite{deyoung2023multi}}  \\ \cline{2-5} 
                                                    & ROUGE-L & \multicolumn{1}{c|}{19.8~{\footnotesize\cite{deyoung2023multi}}}    & \multicolumn{1}{c|}{\textbf{21.1}~{\footnotesize\cite{deyoung2023multi}}}    & \multicolumn{1}{c|}{19.6~{\footnotesize\cite{deyoung2023multi}}} \\ \hline
\end{tabular}
\end{center}
\end{table*}

ROUGE-1 measures the overlap of unigrams between the generated summary and the reference summary. ROUGE-2 measures the overlap of bigrams between the two summaries~\cite{lin2004rouge}. 
 % On the other hand, as discussed earlier, ROUGE-L is a variant of ROUGE that utilizes the Longest Common Subsequence (LCS) instead of n-grams. It calculates the length of the longest sequence of words that appears in both the generated summary and the reference summary, divided by the total number of words in the reference summary~\cite{lin2004rouge}.
In this research, we use ROUGE-1, ROUGE-2, and ROUGE-L to evaluate the quality of the generated summaries. 
% These metrics are commonly used in summarization tasks because they provide a quantitative assessment of the generated summary's similarity to the reference summary~\cite{ma2020multi}. They also provide a simple and efficient way to compare different summarization models or algorithms. 
The importance of these metrics lies in their ability to accurately measure the effectiveness of the summarization process, allowing researchers to evaluate and compare different summarization techniques in a meaningful way. Therefore, ROUGE-1, ROUGE-2, and ROUGE-L can be considered as essential evaluation metrics for summarization tasks~\cite{afsharizadeh2022survey, ma2020multi}.

\section{Results and Discussion}
This section presents the findings of our study, which compares the performance of state-of-the-art models on different datasets from various domains (Table~\ref{tab:ROUGE-scores}). Specifically, we report on the performance of these models using ROUGE scores. We also examine the impact of dataset characteristics, such as the number of documents and documents per cluster, on the performance of the models  (Table~\ref{tab:dataset}).

Our evaluation of the PRIMERA model revealed that it performs well on the Multi-News dataset, achieving a ROUGE-1 score of $42.0$. However, its performance decreases on other datasets such as Multi-XScience, Rotten Tomatoes, and WikiSum, with ROUGE-1 scores of $29.1$, $25.4$, and $28.0$, respectively. Additionally, our results show that PRIMERA's performance is the lowest on the BigSurvey-MDS dataset, with a ROUGE-1 score of $23.9$.

In comparison, the PEGASUS model's performance is relatively consistent across different domains. It achieved ROUGE-1 scores of $32.0$, $27.4$, and $27.6$ on Multi-News, Rotten Tomatoes, and Multi-XScience datasets, respectively. Although these scores are lower than PRIMERA's scores, the difference is not significant. However, PEGASUS's performance on WikiSum dataset is also lower, with a ROUGE-1 score of $24.6$.
It is noteworthy that PEGASUS achieved a higher ROUGE-1 score of $38.9$ on BigSurvey-MDS dataset, indicating that it is relatively consistent across different domains.
In the MS\textasciicircum{}2 benchmark, which is designed to evaluate MDS models in the biomedical domain, LED outperformed PRIMERA and PEGASUS in terms of ROUGE scores. PRIMERA obtained a ROUGE-1 score of $12.8$, while PEGASUS achieved $12.7$. In contrast, LED obtained a significantly higher ROUGE-1 score of $25.8$.
On the other hand, while PRIMERA is a state-of-the-art model and outperforms PEGASUS in some cases, its performance decreases on certain domains. This could be attributed to the fact that PRIMERA was trained on NewSHead corpus which is also in the news domain, and therefore may not be generalized well for other domains.

% Our analysis suggests that the number of documents and documents per cluster have a significant impact on the performance of these models. Specifically, as the number of documents per cluster increases, the performance of the models tends to decrease. For instance, Multi-News and Multi-XSci datasets have a relatively low number of documents per cluster (2.8 and 4.4, respectively), while WikiSum has a much higher number of documents per cluster (40.0). This may explain the relatively poor performance of the models on the WikiSum dataset.
In order to analyse the impact of the number of documents and documents per cluster on model performance, it is important to compare models within the same domain since varying domains can also affect performance. We compared the performance of models on Multi-XScience and BigSurvey-MDS datasets, both of which are in the scientific publication domain. We found that PEGASUS and LED models performed better than PRIMERA on the BigSurvey-MDS dataset. It is worth noting that Multi-XScience dataset had a relatively low number of documents per cluster, while BigSurvey-MDS had the highest number ($2.8$ and $61.4$, respectively). This could explain the relatively better scores on the BigSurvey-MDS dataset.
Overall, our evaluation suggests that the performance of state-of-the-art models for MDS can vary across different domains and based on the dataset characteristics. 
%Future research could explore ways to improve the performance of these models on datasets with higher documents per cluster.

\section{Conclusion and Future Directions}
MDS is an important and rapidly growing research area with the potential to transform the way we process and comprehend large amounts of information. % Significant progress has been made in the field through the use of traditional summarization techniques, deep learning approaches, and evaluation metrics. 
However, there are several challenges to be addressed in MDS, including dealing with diverse document sets, handling redundant information, and ensuring coherence and consistency in the final summary.
Future research should focus on overcoming these challenges and improving the quality of summarization techniques. One promising direction is to improve the generalization of state-of-the-art models across different domains, as they tend to perform differently based on the number of documents per cluster and domain changes. Furthermore, integrating additional factors like sentiment into MDS can capture other valuable dimensions of information that have yet to be extensively explored.
% Additionally, incorporating other factors such as sentiment into MDS can capture other useful aspects of information that remain largely unexplored.
%
In summary, the field of MDS holds great potential for improving information processing, and continued research efforts are necessary to address the challenges and to improve the quality of summarization.

\bibliography{bibliography}{}
\bibliographystyle{IEEEtranN}

\end{document}